\RequirePackage{fix-cm}
\documentclass[twocolumn]{svjour3}          % twocolumn
\smartqed  % flush right qed marks, e.g. at end of proof
\usepackage{graphicx}
\usepackage{booktabs}
\usepackage{multirow}
\usepackage{amsmath}
\usepackage{color}
\usepackage{algorithm}  
\usepackage{algorithmicx}    
\usepackage{algpseudocode} 
\usepackage{balance}
\usepackage[numbers]{natbib}
\usepackage{caption}
\usepackage{indentfirst}
\begin{document}

\title{DGST : Discriminator Guided Scene Text detector%\thanks{Grants or other notes
}


\author{Jinyuan Zhao \and Yanna Wang \and Baihua Xiao \and Cunzhao Shi \and Fuxi Jia \and Chunheng Wang %etc.
}
%\authorrunning{Jinyuan Zhao \and Yanna Wang \and Baihua Xiao \and Cunzhao Shi \and Fuxi Jia \and Chunheng Wang} % if too long for running head

\institute{Jinyuan Zhao \at
              The State Key Laboratory of Management and Control for Complex Systems, Institute of Automation Chinese Academy of Sciences, Beijing 100190, China \\
              University of Chinese Academy of Sciences, China \\
              Tel.: +86-10-82544488\\
              Fax: +86-10-62650820\\
              \email{zhaojinyuan2016@ia.ac.cn}           %  \\
}

\date{Received: date / Accepted: date}

\maketitle

\begin{abstract}
Scene text detection task has attracted considerable attention in computer vision because of its wide application. In recent years, many researchers have introduced methods of semantic segmentation into the task of scene text detection, and achieved promising results. This paper proposes a detector framework based on the conditional generative adversarial networks to improve the segmentation effect of scene text detection, called DGST (Discriminator Guided Scene Text detector). Instead of binary text score maps generated by some existing semantic segmentation based methods, we generate a multi-scale soft text score map with more information to represent the text position more reasonably, and solve the problem of text pixel adhesion in the process of text extraction. Experiments on standard datasets demonstrate that the proposed DGST brings noticeable gain and outperforms state-of-the-art methods. Specifically, it achieves an F-measure of 87\% on ICDAR 2015 dataset.

\end{abstract}
\section{Introduction}
Text detection in natural scenes has attracted more and more attention in the field of computer vision due to its wide application in various natural scene understanding tasks, such as scene location, automatic driving, text analysis, etc. 

In recent years, a lot of scene text detection technologies have emerged, and have achieved good performance in various competitions and public datasets. However, there are still many challenges in the task of scene text detection, such as changing fonts, languages, complex lighting and background conditions, confusion of similar patterns and logos, etc. Figure 1 shows sample images of some scene text detection tasks.
\begin{figure}[t]
\centering
\includegraphics[scale=0.12]{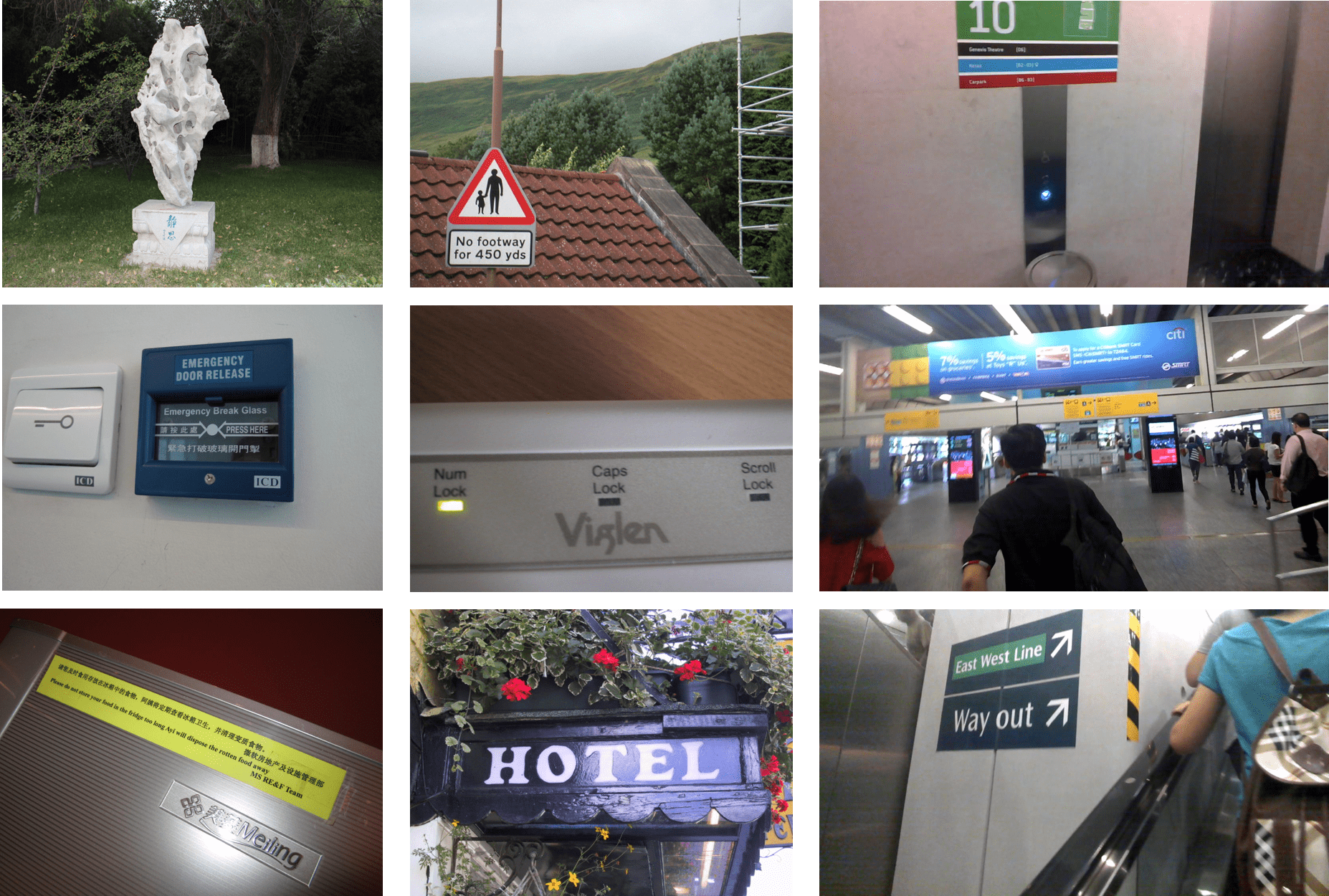}
\caption{Some scene text image examples taken from public datasets.}
\end{figure} 

Existing scene text detection frameworks are mainly inspired by general object detection methods and semantics segmentation methods. The methods based on general object detection usually consist of two stages: RPN network extracts candidate text regions, and classification network sorts the features from the RPN network and obtains the final text position. Semantic segmentation based methods usually treat text as a special segmentation instance, hoping to directly distinguish it from the background in the segmentation results. These methods are called one-stage methods. Compared with two-stage methods, one-stage methods are more intuitive and concise, but still, have the following problems:

\textbf{ Imprecise segmentation labels:} Traditional one-stage methods often train the networks to get a binary text score map. However, due to the diversity of text distribution in scene text images, many annotated text boxes will contain some background pixels. When text pixels are used as a target to conduct pixel-level instance segmentation, these background pixels may cause the problem of learning confusion and reduce the effect of training.

\textbf{Multitask learning problem:} Some classic one-stage methods, such as EAST~\citep{east}, adopt the strategy of obtaining text score map and features required by regression task from the same convolution network. However, regression information, as a distance measure, cannot share features extracted from the CNN network well with text score map based on graph features, and its performance is slightly weaker than that of the two-stage detector.

In this paper, we propose the discriminator guided scene text detector (DGST) to address the above problems and improve the performance of one-stage text detectors. We introduce the framework of conditional generative adversarial networks, which is popular in image generation task recently. Text detection task is transformed into related segmentation image generation tasks. A discriminator is used to automatically adjust the losses in training process and generate a satisfactory text score map. At the same time, we design the soft-text-score map to strengthen the center position of text boxes and weaken the influence of edge pixels on the detection results, so as to eliminate the interference of background pixels and avoid learning confusion in the learning process. The final detection results can be obtained by combining the soft-text-score maps of different shrink factors. We evaluated our method on ICDAR2013~\citep{icdar2013}, ICDAR2015~\citep{icdar2015}, ICDAR2017~\citep{icdar2017} and MSRA-TD500~\citep{MSRA} datasets. Among them, the F-measure of our method reaches 87\% on ICDAR2015~\citep{icdar2015} and 74.3\% on ICDAR2017~\citep{icdar2017}.

Our pipeline is shown in Fig.2. The main contributions of this paper are three-fold:

 $\bullet$ We introduce the framework of generative adversarial networks into the task of scene text detection and design a suitable structure for it.
 
 $\bullet$ We redefine the representation of text area and non-text area in the framework of semantic segmentation, and solve the learning confusion caused by background pixels.
 
 $\bullet$ Extensive experiments demonstrate the state-of-the-art performance of the proposed method on several benchmark datasets.

\section{Related Works}
With the development of computer technology and the popularization of deep-learning methods, detectors based on neural network framework have shown excellent performance in scene text detection tasks, which makes text detection enter a new era of deep-learning methods.

Many works have been done on scene text detection in recent years. These methods can be divided into two branches: one branch is based on general object detection methods such as SSD~\citep{ssd}, YOLO~\citep{yolo}, and Faster RCNN~\citep{faster1}. TextBoxes++~\citep{Textboxes++} modifies anchors and kernels of SSD~\citep{ssd} to enable the detector to process texts of large aspect ratio in scene images. RRPN~\citep{rrpn} changes the aspect ratio of anchor in Faster RCNN~\citep{faster1} and adds rotation anchors to support scene text detection with arbitrary orientation. CTPN~\citep{CTPN} further analyses the characteristics of text, optimizes RPN in Faster RCNN~\citep{faster1} to extract candidate box and merge many small candidate boxes into the final text prediction box, so as to solve the problem of text line detection of arbitrary length. These text detectors take words or text lines as a special object and add subsequent classifiers to filter text areas in convolution features. Usually, these methods need to add NMS to get the final text location.

Another branch is based on semantic segmentation, which regards scene text detection as a special semantics segmentation task. Zhang et al.\citep{zhangfcn} uses FCN to estimate text blocks and MSER to extract candidate characters.  EAST~\citep{east} adopts the idea of FCN, and predicts the location, scale, and orientation of text with a single model and multiple loss functions (multi-task training). PSENET~\citep{PSENET} uses semantic segmentation to classify text at the pixel level, which makes the modeling of curved text simpler and uses kernels to separate close text blocks. CRAFT~\citep{CRAFT} takes the affinity between characters and characters itself as different target instances to generate scoring graphs and detects text at the character level. These methods hoping to get a binary text score graph and extract texts in the image as segmentation instances. The final text position can be obtained by analyzing the text score map. Compared with the two-stage methods, these methods have more intuitive ideas and simpler network structure.
%, but sometimes they have a slight loss in the accuracy of detection.

These methods above have achieved excellent performance on standard benchmarks. However, as illustrated in Fig. 3(a), the problem of imprecise segmentation labels has not been well solved, especially for semantically segmented detectors, the background pixels in the annotation boxes will affect the classification results, which leads to the deviation of the final results. Meanwhile, many methods need to learn multiple tasks at the same time, such as classification, regression, and text score-map generation, which makes the network structure and inference more complex.

Some semantics-based detectors have explored the text representation and improved the previous score map labeling methods: PixelLink~\citep{Pixellink} first transforms text detection into a pure segmentation problem by linking pixels within the same instance of eight-directions and then extracts the text boundary box directly from the segmentation without location regression. PSENet~\citep{PSENET} finds text kernels of different scales and proposes a progressive scaling expansion algorithm to accurately separate cohesive text instances. Textfield~\citep{xu2019textfield} uses the direction field which encodes both binary text mask and direction information facilitating the subsequent text grouping process.

With the emergence of deep-learning techniques, the research on the direction of generative image modeling has made significant progress~\citep{GAN1,GAN2,GAN3}. ~\citep{taigman2016unsupervised} uses the conditional GANs to translate a rendering image to a real image.  An unsupervised image-to-image translation framework based on shared latent space is proposed in ~\citep{liu2017unsupervised}. More recently, CycleGAN~\citep{zhu2017unpaired} and its variants~\citep{yi2017dualgan, kim2017learning} have achieved impressive image translation by using cycle-consistency loss. \citep{hoffman2018cycada} proposes a cycle-consistent adversarial model that is applicable at both pixel and feature levels.

Inspired by the above methods, in this paper, we use the generative adversarial networks framework and design more reasonable soft-text-score map to get more accurate semantic segmentation results and use connected components analysis to replace the traditional NMS process. This not only avoids the learning confusion caused by imprecise labels but also makes the whole network training process become a single task learning process, which is more concise and intuitive.

\section{METHODOLOGY}
\begin{figure*}[!t]
\centering
\includegraphics[scale=0.15]{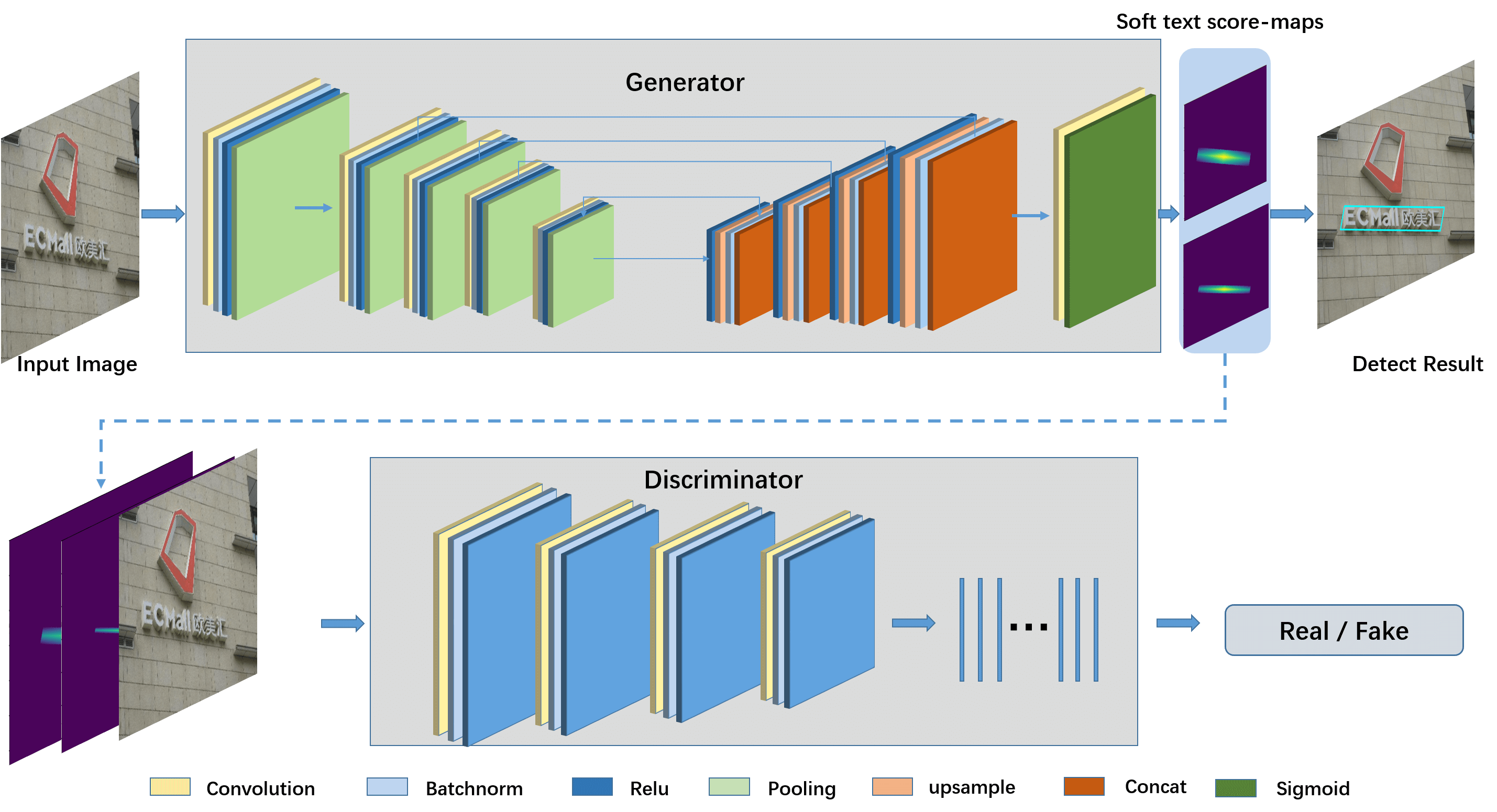}
\caption{Overview of the proposed DGST.}
\end{figure*} 
Fig.2 shows the flowchart of the proposed method for scene text detection, which is a one-stage detector. In the training process, the generator and discriminator learn alternately, so that the generator finally converts the input scene image into the corresponding soft-text-score map. This eliminates intermediate steps such as candidate proposal, thresholding, and NMS on predicted geometric shapes. The post-processing steps only include connected components analyses of the text score map. The detector is named as DGST since it is a Discriminator Guided Scene Text detector.

\subsection{Label Generation}
Some classical one-stage detectors usually generate a binary text score map, such as EAST~\citep{east}, PSENET\citep{PSENET} and Pixel-Link~\citep{Pixellink}. However, this labeling method has the drawbacks mentioned in Section 1. When text feature extraction is regarded as a semantic segmentation task to classify the input image at the pixel level, the background pixels in the ground-truth boxes will interfere with the learning of text features. Some of these methods try to shrink the annotation boxes more tightly to reduce the background pixels, as shown in Fig.3~(a). However, such a rigid shrinkage can not accurately adjust the labeling of each box, and the text edges and background pixels can not be well distinguished, which makes the final text box position deviate from the desired result. CRAFT~\citep{CRAFT} method divides the text line annotation into single character annotation results and measures the Gauss distance on each character to get the text score map, which further weakens the influence of background noise on text feature extraction, but the conversion from word-level annotation to character-level annotation introduces additional complex work.
\begin{figure}[!t]
\centering
\includegraphics[scale=0.15]{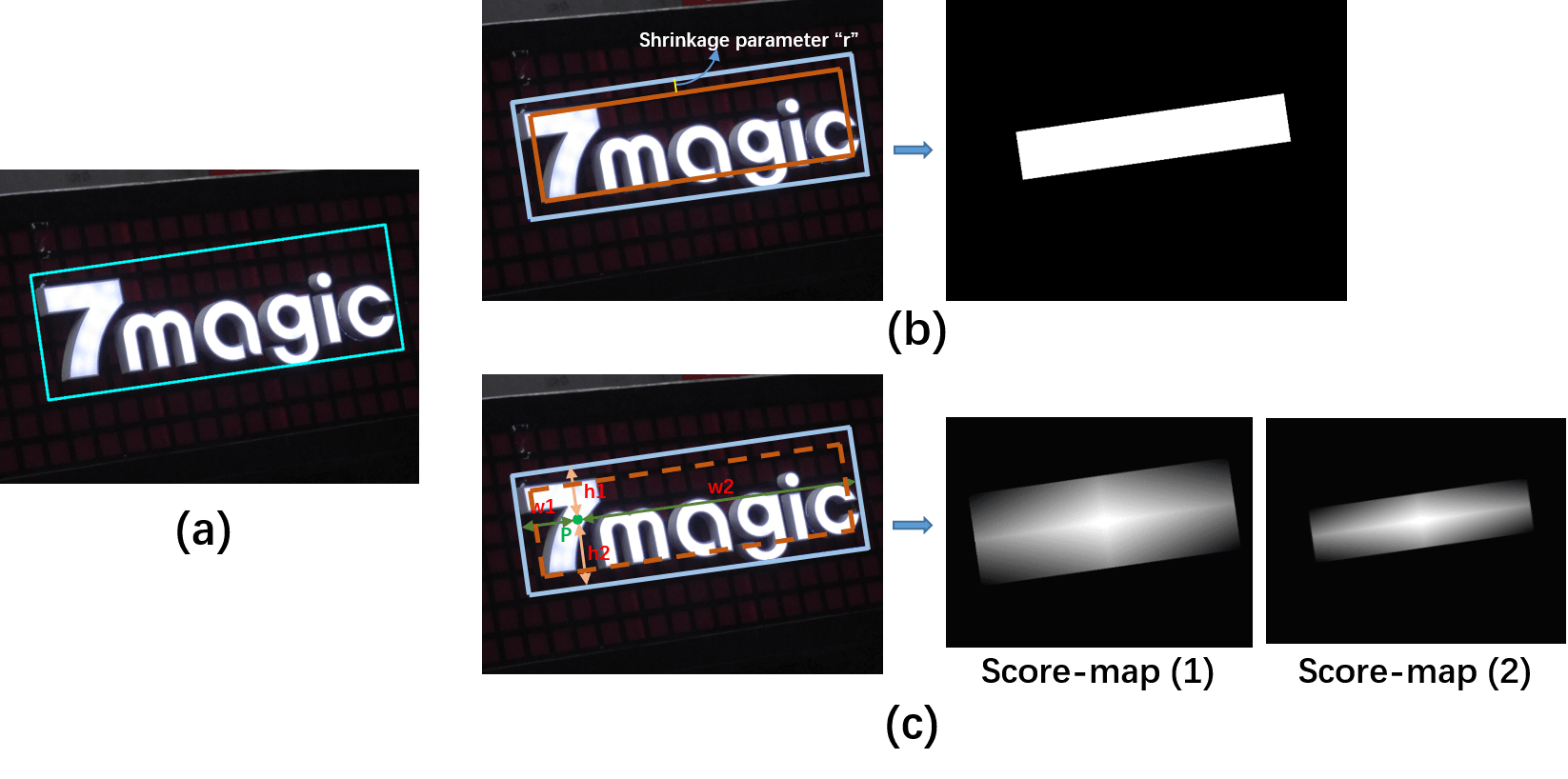}
\caption{Diagrams of different text score map annotation methods. (a) The labeling method used by EAST~\citep{east}. (b) The labeling method proposed by this paper.}
\end{figure} 

In this paper, inspired by the above methods, we propose a method to generate text score maps based on distance pairs between the pixels in the annotation box and the corresponding boundaries. We compare the distance between the pixels in the annotation box and the corresponding boundary in horizontal and vertical directions, highlighting the central position of the text line, and weakening the weight of the pixels on the edge, which are easily confused with the background. For a point $ (x, y) $ in the input image, its intensity value $ P $ in soft-text-score map can be calculated by the following formula:
\begin{equation}
P_{(x,y)}=\left\{
\begin{aligned}
 &\frac{1}{2} \times (D_{w}+D_{h}) & , & (x,y)\in T_{i}, \\
 & 0 & , & (x,y) \in background.
\end{aligned}
\right.
\end{equation}

\begin{equation}
D_{w}=1-\dfrac{\left|w_{i2}-w_{i1}\right|}{w_{i}}
\end{equation}
\begin{equation}
D_{h}=1-\dfrac{\left|h_{i2}-h_{i1}\right|}{h_{i}}
\end{equation}

Where we use set $ T $ to represent all annotated text boxes, $ w_{i} $ and $ h_{i} $ represent the width and height of the i-th text box, respectively. $ w_{i1} $,$ w_{i2} $, $ h_{i1} $, $ h_{i2} $ denote the distance of point $ (x, y) $ to each edge. We use the everage of $ D_{w} $ and $ D_{h} $ to calculate the gray value $ P $, which decreases from the center line in the horizontal and vertical direction to the edge points in every text box. An intuitive display is shown in Fig.3~(b). \iffalse For the text lines labeled with polygons, $ w_{i} $ and $ h_{i} $ represent the distance from the point $ (x, y) $ to the boundary corresponding to the horizontal and vertical directions, respectively.\fi

The values of all the pixels are between [0,1]. In order to solve the problem that it is difficult to deal with cohesive text blocks in post-processing, we generate two different levels of score maps for the same input image. The pixel values in the two score maps are calculated in exactly the same way. The difference is that the text box in score map (2) is contracted in the way shown in Fig.3~(a) so that there is a greater gap between the text boxes (as shown in the dotted line box in Fig.3~(b)). In our experiment, the contraction factor is 0.2. 

\subsection{Network Design}

\begin{figure*}[!t]
\centering
\includegraphics[scale=0.2]{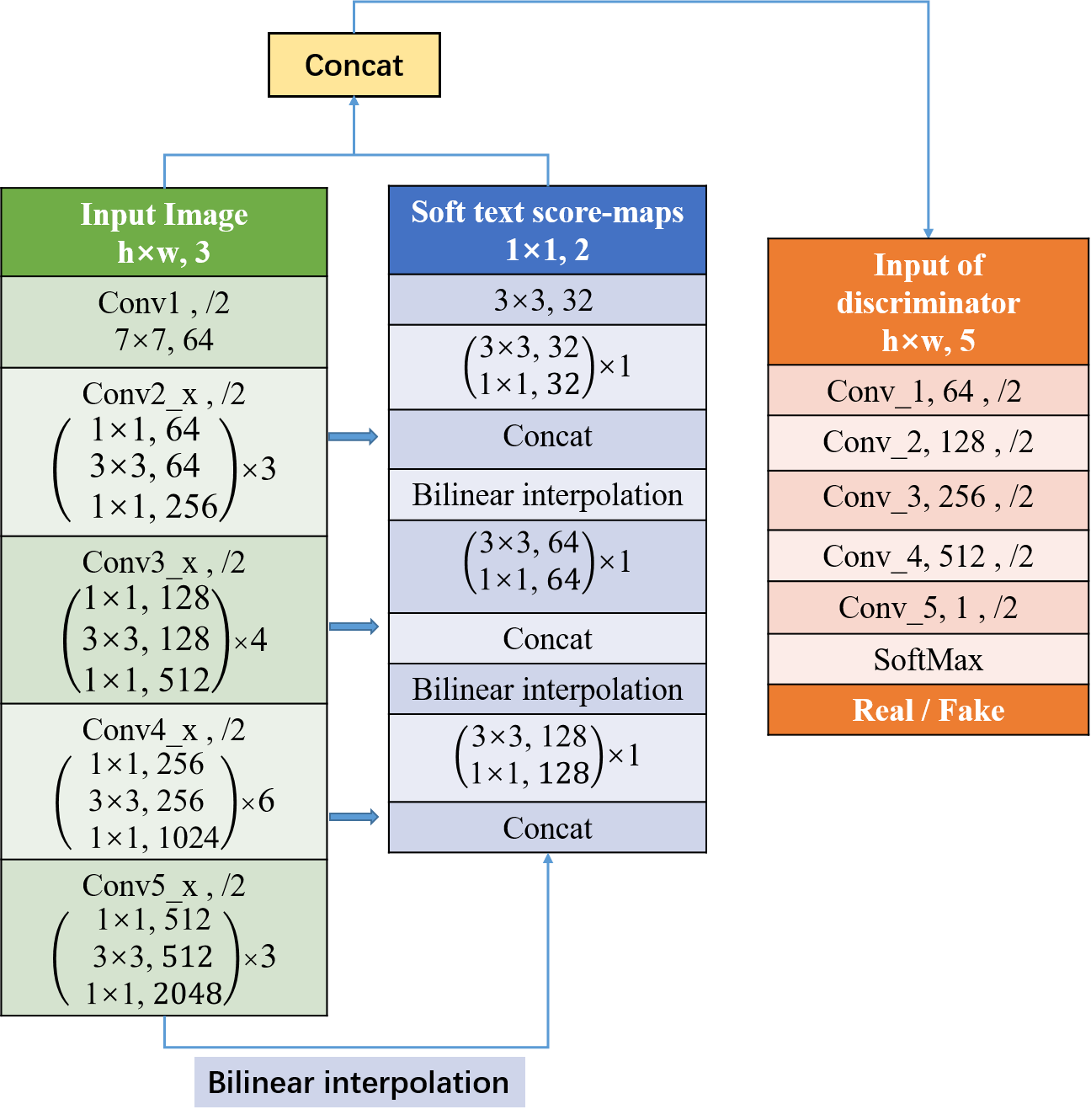}
\caption{Network structure of the proposed method. The upsampling operation is done through bilinear interpolation directly. Feature maps from different stages are fused through a cascade of upsampling and add operations. $ (n \times n, X) $ denotes a convolution layer with $ X $ convolution kernels of size $ n \times n $.}
\end{figure*} 
\subsubsection{Generator and discriminator}
We use U-shaped network structure to fuse the feature in down-sampling and up-sampling step by step. This strategy has been validated in many previous scene text detection methods such as~\citep{CRAFT,east} and~\citep{Pixellink}. We use ResNet-50~\citep{ResNet} as the backbone of DGST, and the feature maps of \{ Conv2\_x, Conv3\_x, Conv4\_x, Conv5\_x  \} are combined by up-sampling. 

From an input image, five levels of the feature maps are combined to generate the final feature maps. With the help of discriminator, our generator outputs a two-channel feature map with the same scale as the input image, representing the soft text score maps under different shrink factors respectively. Therefore, the feature extraction task of traditional text detection is transformed into a feature image generation task.

Combining the original picture with the corresponding text score maps of different shrink factors as the input of the discriminator, the discriminator determines whether the input text score map is a labeled ground truth image or an imitation of the discriminator.

A more detailed network structure is shown in Fig.4. We use bilinear interpolation instead of deconvolution to avoid the chessboard effect. The green and blue tables in the figure are the network structure of the generator's feature extraction and fusion phase respectively, and the orange table is the network structure of our discriminator.

\subsubsection{Loss function}
Traditional GAN images are trained alternately by game learning of generators and discriminators. Their loss functions are as follows:
\begin{equation}
\qquad \qquad \arg \min\limits_{G} \max\limits_{D}L_{cGANs}(G,D)
\end{equation}  

In order to obtain a more accurate score map, we use the following two measures to further strengthen the generator on the basis of the traditional GAN structure:

1. cGAN is used instead of traditional GAN structure. Input pictures are added as a restriction, so that the output of the generator is restricted by input pictures, and more reasonable result images can be obtained. The loss function is as follows:
 \begin{equation}
 \begin{aligned}
 & L_{cGANs}(G,D)  = \\
 & \quad E_{x,y}[\log D(x,y)]+E_{x,z}[\log(1-D(x,G(x,z)) )]
 \end{aligned}
 \end{equation}
 
2. On the basis of GAN loss, the traditional loss function L2-loss is introduced to optimize the predicted text score map, which makes the generated text score map not only deceive the discriminator but also perform better in the sense of traditional loss.
\begin{equation}
\qquad \qquad L_{L2}(G)=E_{x,y,z}[\parallel y-G(x,z) \parallel_{2}]
\end{equation}
The final loss function is as follows:
\begin{equation}
\qquad G^{*}=\arg \min\limits_{G} \max\limits_{D}L_{cGANs}(G,D)+\lambda L_{L2}(G)
\end{equation}

\begin{figure*}[!t]
\centering
\includegraphics[scale=0.2]{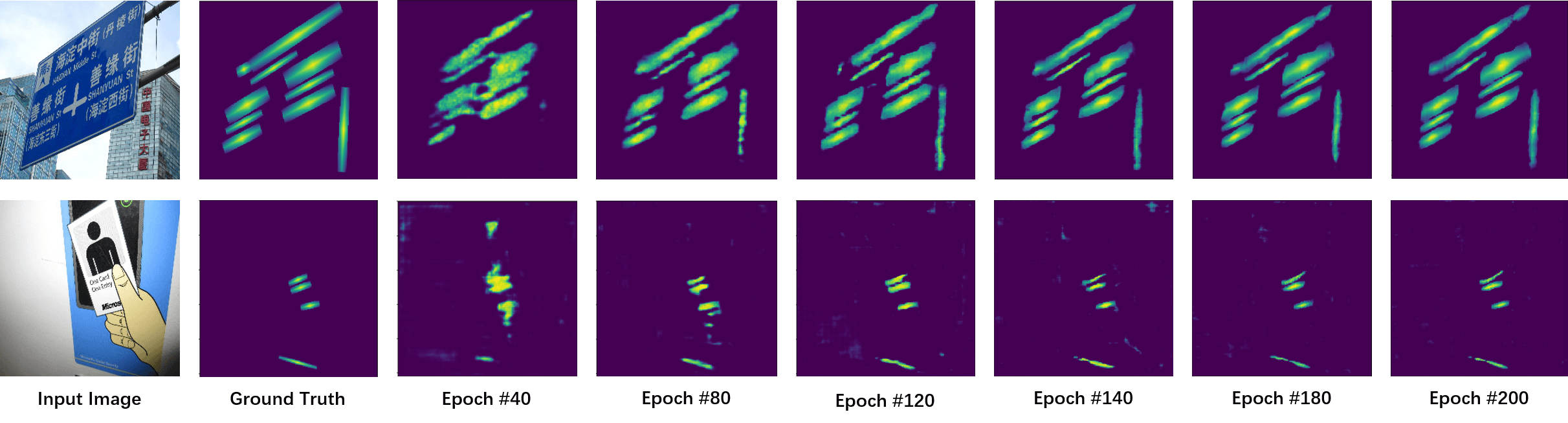}
\caption{Text score maps generated in different epochs (contraction factor is 0).}
\end{figure*} 

Fig.5 shows the text scoremap~(1) generated by our DTDR in different epochs. As the number of iterations increases, the text score map generated by our generator can continuously approximate the given GT and further filter out the noise interference in the background.

\subsection{Text boxes extraction}
Fig.6 shows the overall flow of our post-processing method. Two text score maps with different shrink factors are obtained from the generator, and the corresponding text boxes in Fig.6~(c) and Fig.6~(d) can be obtained by directly analyzing the connected components of score maps in Fig.6~(b). It can be seen that there is a cohesion problem in non-shrinking score map, and the shrinking score map can better extract text box spacing information, but it will lose some text information. 

\begin{figure*}[!t]
\centering
\includegraphics[scale=0.13]{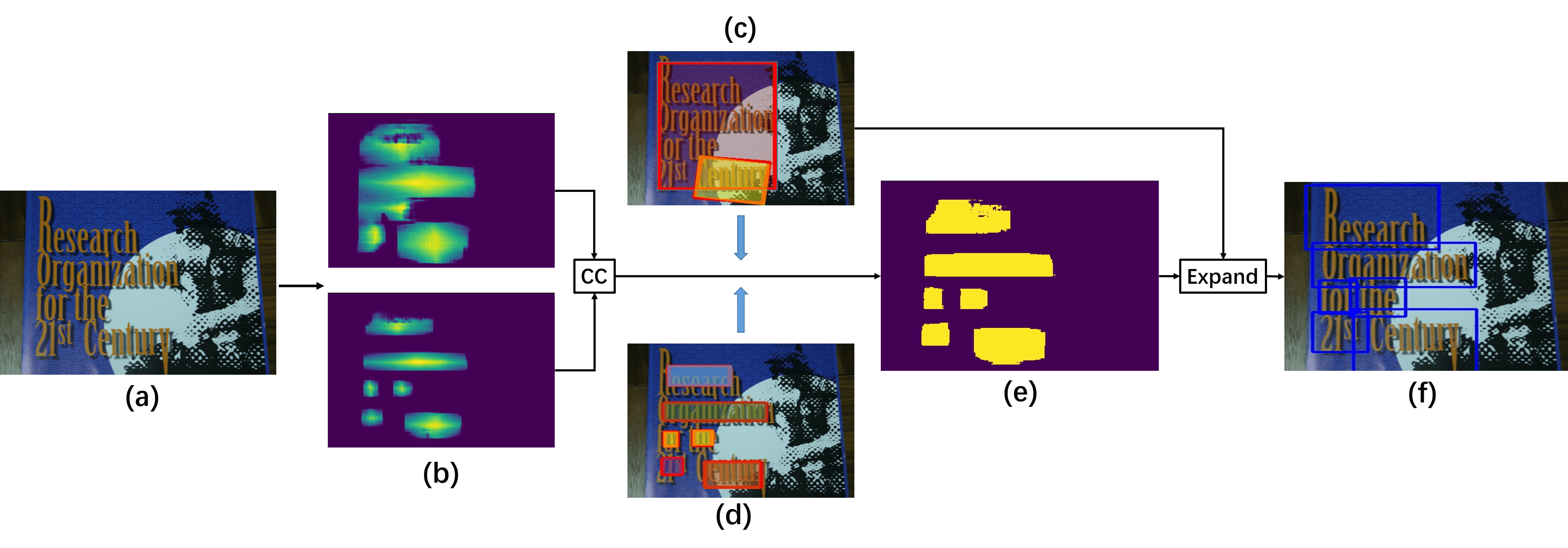}
\caption{An illustration of extracting text location information from score maps. (a) Original input image. (b) Score maps of different contraction factors generated by DGST. (c) (d) The connected component analysis results of images in (b). (e) The binary result obtained by fusing the two maps in (b). (e) The final result of text detection.}
\end{figure*} 

Therefore, we combine the two score maps from the generator to get a more complete image as shown in Fig.6~(e), and expand the text boxes from Fig.4~(e) under the constraint of the text boxes in Fig.6~(c), so that the edge can surround the whole text area completely. The final text box position is shown in Fig.6~(f). More specific processes are shown in algorithm 1:
\begin{algorithm}
        \caption{Text boxes extraction }
        \begin{algorithmic}[1] %每行显示行号
            \Require The text score map $M_{score_1}$ and $M_{score_2}$ with different shrink factors
            \Ensure The set of text boxes $ T_{Q} $
            	\State $B_{s1}$ = threshold ($M_{score_1}$, $ t $)  
                \State $B_{s2}$ = threshold ($M_{score_2}$, $ t $)  
                \State $  T_{Q1} $ = ConnectedComponents($B_{s1}$)
%                \State $  T_{Q2} $ = connectedComponents($B_{s2}$)
                \State Combination score map $M_{score}$ =( $M_{score_1}$ + $M_{score_2}$)
                \State $B_{s}$ = threshold ($M_{score}$, $ t $)  
                \For  {$(x,y)$ in $B_{s}$}
                      	\If  {$P(x,y)$ in $B_{s2}$ $==0$}
                      		\State $B_{s}(x,y)=0 $
                      	\EndIf
                \EndFor
                \State $  T_{Q'} $ = connectedComponents($B_{s}$)
                \For  {$Q$ in $ T_{Q'} $}
                    \For  {$Q1$ in $  T_{Q1} $}
                         	\If  {$Q$ is surrounded by $Q1$}
                         		\State Expanding $Q$ until it coincides with an edge of $Q1$.
                         	\EndIf
                    \EndFor
                    \State  $T_{Q_temp}\gets Q $
                \EndFor
                \State  $T_{Q}\gets T_{Q_temp} $
          %  \EndFunction 
        \end{algorithmic}
\end{algorithm}

$M_{score_1}$ and $M_{score_2}$ denote the two score maps with different shrink factors. $B_{s1}$ and $B_{s2} $ denote the binary image of $M_{score_1}$ and $M_{score_2}$ respectively. Here we introduce a paramenter $t$ to threshold the score maps. we choose $t=0.25$ in our experiments. Relevant operations such as thresholding and connected components analysis can be implemented with the correlation functions provided by OpenCV.

\section{EXPERIMENTS}

To verify the effectiveness of the proposed method in scene text detection task, we compare the performance of DGST with existing methods on several standard benchmarks: ICDAR 13, ICDAR 15, ICDAR 17 and MSRA-TD500. The experimental results show that we have achieved on better or comparable results than state-of-the-art methods.

\subsection{Datasets}

\textbf{ICDAR2013} (IC13)~\citep{icdar2013} was released during the ICDAR 2013 Robust Reading Competition for focused scene text detection. ICDAR2013 dataset is a subset of ICDAR2011 dataset. The number of images of ICDAR2013 dataset is 462, which is comprised of 229 images for the training set and 233 images for the test set. This dataset only contains texts in English. The annotations are at word-level using rectangular boxes.

\textbf{ICDAR2015} (IC15)~\citep{icdar2015} was introduced in the ICDAR 2015 Robust Reading Competition for incidental scene text detection. 1,500 of the images have been made publicly available, split between a training set of 1, 000 images and a test set of 500, both with texts in English. The annotations are at the word level using quadrilateral boxes.

\textbf{ICDAR2017} (IC17)~\citep{icdar2017} was introduced in the ICDAR 2017 robust reading challenge on multi-lingual scene text detection, consisting of 9000 training images and 9000 testing images. The dataset is composed of widely variable scene images which contain text of one or more of 9 languages representing 6 different scripts. The number of images per script is equal. The text regions in IC17 are annotated by the 4 vertices of quadrilaterals, as in ICDAR2015.

\textbf{MSRA-TD500} (TD500)~\citep{MSRA} contains 500 natural images, which are split into 300 training images and 200 testing images, collected both indoors and outdoors using a pocket camera. The images contain English and Chinese scripts. Text regions are annotated by rotated rectangles.

\subsection{Evaluation protocol}

We use standard evaluation protocol to measure the performance of detectors in terms of precision, recall, and f-measure. They are defined as follows:
\begin{equation}
\qquad \qquad Precision=\frac{TP}{TP + FP} 
\end{equation}
\begin{equation}
\qquad \qquad Recall=\frac{TP}{TP + FN}
\end{equation}
\begin{equation}
\qquad  F-measure=\frac{2 \times Recall \times Precision}{Recall + Precision}
\end{equation}
 where $ TP $, $ FP $, $ FN $ denote the True Positive, False Positive and False Negative values, respectively. For the detected text instance T, if the IOU is greater than the given threshold when T intersects a ground truth text instance (usually set to 0.5), then the text instance T is considered to be the correct detection. Because of the trade-off between recall and precision, F-measure is a common compromised measurement for performance evaluation.
 
\subsection{Implementation details}

The DGST is implemented in Pytorch framework and run on a server with 2.10GHz CPU, GTX 1080Ti GPU, and Ubuntu 64-bit OS. The layers of our generator are initialized with the backbone models (ResNet-50) pretrained on ImageNet~\citep{imgnet}. We choose minibatch SGD and apply the  Adam solver~\citep{adam} with learning rate 0.0002. 

When experimenting on a specific data set, the training set is augmented by existing training samples. The specific ways of expansion are as follows: (1) Each image is randomly scaled between 640-2560 in length or width, and the original aspect ratio is maintained. (2) Rotate each training image randomly at four angles [0,90,180,270]. (3) Random crop 640$ \times $640 regions in the scaled image (pure background area does not exceed 30\% of the total sample number). For the other methods in Tab.1,2,3 and 4, we directly use the experimental results shown in the original paper to compare with our results.

%\subsection{Advantages of GAN framework}
\subsection{Ablation Experiments}
We use the evaluation indicators in Section 4.2 and compare different network structures on the ICDAR15 test set. Table~\ref{Ablation Experiments} summarizes the experimental results.

Our baseline is a U-net structure with ResNet50 as the backbone network, and uses cross-entropy loss to train a binary text score map.  On this basis, we compare the effects of soft text representation and the discriminator training strategy on detector performance.

\begin{table}[htbp]
\renewcommand\arraystretch{1.2}
  \small
  \centering
  \caption{Results on the ICDAR15 test set under different model configurations and training strategies.}
             \setlength{\abovetopsep}{2ex}
             \setlength{\belowrulesep}{0pt}
             \setlength{\aboverulesep}{0pt}
             \setlength{\tabcolsep}{1.2mm}
    \begin{tabular}{|c|c|c|c|ccc|}
    \toprule
    \multicolumn{4}{|c|}{\textbf{Method}} & \textbf{Recall} & \textbf{Precision} & \textbf{F-score} \\
    \midrule
    \multicolumn{4}{|c|}{Baseline} & 82.1  & 83.4  & 82.7 \\
    \multicolumn{4}{|c|}{Baseline+soft text score map} & 84.2  & 86.9  & 85.5 \\
    \multicolumn{4}{|c|}{Baseline+GAN loss} & 82.9  & 85.5  & 84.2 \\
    \midrule
    \multicolumn{4}{|c|}{\textbf{DGST}} & \textbf{84.7} & \textbf{89.6} & \textbf{87.1} \\
    \bottomrule
    \end{tabular}%
  \label{Ablation Experiments}%
\end{table}%

In our ablation experiment, except for the differences mentioned in the first column of the Table~\ref{Ablation Experiments}, the model structure and training strategy of other experimental links are exactly the same as the baseline. Among them, DGST is our final detector structure, which combines two strategies of soft text score map and Gan loss on the basis of baseline. 

From the Table~\ref {Ablation Experiments}, we can see that using the soft text score map proposed in Section 3 instead of the traditional binary text score map can significantly improve the detection results. For the pixel level segmentation task, more abundant classification information can distinguish the text pixel and non text pixel information in the annotation box, which can significantly improve the classification accuracy of the final image pixel, so as to get more accurate detection results. In the meantime, similar to many semantic segmentation tasks, we use the conditional generative adversarial training strategy instead of traditional cross-entropy loss to train the generator, so that the classification results can continuously approximate the designed ground truth images, and also can improve the final pixel classification accuracy. Our final detector, DGST, combines the advantages of these two improvements and achieves the optimal effect on the test set.

\subsection{Compare with  Other Methods}
\begin{table}[htbp]
  \renewcommand\arraystretch{1.2}
  \small
  \centering
  \caption{Comparison with other results on ICDAR 2013.}
           \setlength{\abovetopsep}{3ex}
           \setlength{\belowrulesep}{0pt}
           \setlength{\aboverulesep}{0pt}
           \setlength{\tabcolsep}{2mm}
    \begin{tabular}{|c|ccc|}
     \toprule
     \multirow{2}[4]{*}{\textbf{Method}} & \multicolumn{3}{c|}{\textbf{IC13}} \\
 \cmidrule{2-4}          & \textbf{Recall} & \textbf{Precision} & \textbf{F-score} \\
     \bottomrule
     Zhang et al.~\citep{zhangfcn} & \multicolumn{1}{c|}{78} & \multicolumn{1}{c|}{88} & 83 \\
     \midrule
     Yao et al.~\citep{Yao} & \multicolumn{1}{c|}{80.2} & \multicolumn{1}{c|}{88.8} & 84.3 \\
     \midrule
     He et al.~\citep{he} & \multicolumn{1}{c|}{81} & \multicolumn{1}{c|}{92} & 86 \\
     \midrule
     R2CNN ~\citep{R2CNN} & \multicolumn{1}{c|}{82.6} & \multicolumn{1}{c|}{93.6} & 87.7 \\
     \midrule
     TextBoxes++~\citep{Textboxes++} & \multicolumn{1}{c|}{86} & \multicolumn{1}{c|}{92} & 89 \\
     \midrule
     Mask TextSpotter~\citep{Mask} & \multicolumn{1}{c|}{88.1} & \multicolumn{1}{c|}{94.1} & 91 \\
     \midrule
     PixelLink~\citep{Pixellink} & \multicolumn{1}{c|}{87.5} & \multicolumn{1}{c|}{88.6} & 88.1 \\
     \midrule
     FOTS~\citep{Fots} & \multicolumn{1}{c|}{-} & \multicolumn{1}{c|}{-} & 87.3 \\
     \midrule
     Lyu et al.~\citep{Lyu} & \multicolumn{1}{c|}{84.4} & \multicolumn{1}{c|}{92} & 88 \\
     \midrule
     CTPN~\citep{CTPN} & \multicolumn{1}{c|}{\textbf{93}} & \multicolumn{1}{c|}{83} & 88 \\
     \midrule
     SSTD ~\citep{SSTD}  & \multicolumn{1}{c|}{88} & \multicolumn{1}{c|}{86} & 87 \\
     \bottomrule
     \textbf{DGST} & \multicolumn{1}{c|}{91.7} & \multicolumn{1}{c|}{\textbf{95.8}} & \multicolumn{1}{c|}{\textbf{93.7}} \\
     \bottomrule
     \end{tabular}%
   \label{tab:addlabel}%
 \end{table}%
 
 \begin{table}[htbp]
  \renewcommand\arraystretch{1.2}
  \small
  \centering
  \caption{Comparison with other results on ICDAR 2015.}
           \setlength{\abovetopsep}{3ex}
           \setlength{\belowrulesep}{0pt}
           \setlength{\aboverulesep}{0pt}
           \setlength{\tabcolsep}{2mm}
     \begin{tabular}{|c|ccc|}
     \toprule
     \multirow{2}[4]{*}{\textbf{Method}} & \multicolumn{3}{c|}{\textbf{IC15}} \\
 \cmidrule{2-4}          & \textbf{Recall} & \textbf{Precision} & \textbf{F-score} \\
     \midrule
     Zhang et al.~\citep{zhangfcn} & \multicolumn{1}{c|}{43} & \multicolumn{1}{c|}{71} & 54 \\
     \midrule
     Yao et al.~\citep{Yao} & \multicolumn{1}{c|}{58.7} & \multicolumn{1}{c|}{72.3} & 64.8 \\
     \midrule
     He et al.~\citep{he} & \multicolumn{1}{c|}{80} & \multicolumn{1}{c|}{82} & 81 \\
     \midrule
     R2CNN~\citep{R2CNN} & \multicolumn{1}{c|}{79.7} & \multicolumn{1}{c|}{85.6} & 82.5 \\
     \midrule
     TextBoxes++~\citep{Textboxes++} & \multicolumn{1}{c|}{78.5} & \multicolumn{1}{c|}{87.8} & 82.9 \\
     \midrule
     Mask TextSpotter~\citep{Mask} & \multicolumn{1}{c|}{81.2} & \multicolumn{1}{c|}{85.8} & 83.4 \\
     \midrule
     PixelLink~\citep{Pixellink} & \multicolumn{1}{c|}{82} & \multicolumn{1}{c|}{85.5} & 83.7 \\
     \midrule
     FOTS~\citep{Fots} & \multicolumn{1}{c|}{82} & \multicolumn{1}{c|}{88.8} & 85.3 \\
     \midrule
     CRAFT~\citep{CRAFT} & \multicolumn{1}{c|}{84.3} & \multicolumn{1}{c|}{\textbf{89.8}} & 86.9 \\
     \midrule
     Lyu et al.~\citep{Lyu} & \multicolumn{1}{c|}{79.7} & \multicolumn{1}{c|}{89.5} & 84.3 \\
     \midrule
     CTPN~\citep{CTPN} & \multicolumn{1}{c|}{52} & \multicolumn{1}{c|}{74} & 61 \\
     \midrule
     SSTD~\citep{SSTD}  & \multicolumn{1}{c|}{73} & \multicolumn{1}{c|}{80} & 77 \\
     \midrule
     \textbf{DGST} & \textbf{84.7} & 89.6  & \textbf{87.1} \\
     \bottomrule
     \end{tabular}%
   \label{tab:addlabel}%
 \end{table}%
 
\begin{table}[htbp]
  \renewcommand\arraystretch{1.2}
  \small
  \centering
  \caption{Comparison with other results on ICDAR 2017.}
           \setlength{\belowrulesep}{0pt}
           \setlength{\aboverulesep}{0pt}
    \begin{tabular}{|c|c|c|c|}
    \toprule
    \multirow{2}[4]{*}{\textbf{Method}} & \multicolumn{3}{c|}{\textbf{IC17}} \\
\cmidrule{2-4}          & \multicolumn{1}{c}{\textbf{Recall}} & \multicolumn{1}{c}{\textbf{Precision}} & \textbf{F-score} \\
    \midrule
    FOTS~\citep{Fots} & 57.5  & 79.5  & 66.7 \\
    \midrule
    CRAFT~\citep{CRAFT} & 68.2  & 80.6  & 73.9 \\
    \midrule
    Lyu et al.~\citep{Lyu} & \textbf{70.6} & 74.3  & 72.4 \\
    \midrule
    \textbf{DGST} & 67.6  & \textbf{82.6} & \textbf{74.3} \\
    \bottomrule
    \end{tabular}%
  \label{tab:addlabel}%
\end{table}%

\begin{table}[htbp]
  \renewcommand\arraystretch{1.2}
  \small
  \centering
  \caption{Comparison with other results on MSRA-TD500.}
           \setlength{\belowrulesep}{0pt}
           \setlength{\aboverulesep}{0pt}
    \begin{tabular}{|c|c|c|c|}
    \toprule
    \multirow{2}[3]{*}{\textbf{Method}} & \multicolumn{3}{c|}{\textbf{MSRA-TD500}} \\
\cmidrule{2-4}          & \multicolumn{1}{c}{\textbf{Recall}} & \multicolumn{1}{c}{\textbf{Precision}} & \textbf{F-score} \\
\bottomrule
    Zhang et al.~\citep{zhangfcn} & 67    & 83    & 74 \\
    \midrule
    Yao et al.~\citep{Yao} & 75.3  & 76.5  & 75.9 \\
    \midrule
    He et al.~\citep{he} & 70    & 77    & 74 \\
    \midrule
    EAST~\citep{east}  & 67.4  & 87.3  & 76.1 \\
    \midrule
    TextSnake~\citep{TextSnake} & 73.9  & 83.2  & 78.3 \\
    \midrule
    PixelLink~\citep{Pixellink} & 73.2  & 83    & 77.8 \\
    \midrule
    CRAFT~\citep{CRAFT} & 78.2  & \textbf{88.2} & 82.9 \\
    \bottomrule
    \textbf{DGST} & \textbf{79.4} & 87.9  & \textbf{83.4} \\
    \bottomrule
    \end{tabular}%
  \label{tab:addlabel}%
\end{table}%

In order to evaluate the effectiveness of the proposed method, we conducted experiments on the datasets mentioned in subsection 4.1. The proposed method is compared with other state-of-the-art detection algorithms in Recall, Precision, and F-score. Table 1, 2, 3 and 4 show the experimental results on IC13, IC15, IC17, and MSRA-500 datasets respectively. From the results in the tables, we can see that our method achieves the state-of-the-art level on the four datasets and performs well in each evaluation index.

\textbf{ICDAR2017:}  IC17 contains a large number of scene text images in different languages. We use the training set and verification set to finetune the model pre-trained on ImageNet, and iterate 200 epochs to get the final detector. When testing the model, we resize the longer side of images in the test set to 2560 and reaches the F-measure of 74.8\%. The specific results are shown in Table 3.

\textbf{ICDAR2015:}  The images in IC15 and IC17 are similar and contain many small text line instances. Therefore, we use the training set of IC15 to finetune the model from IC17 for 80 epochs, so as to achieve better detection results. For testing, we resized the image to 2240 on the long side for a single scale test,  and the final F-measure was 87.1\%. The specific results are shown in Table 2.

\textbf{ICDAR2013:} Similar to IC15, IC13 also finetune the model from IC17 to get a better detector. Because of the large area of the text area in the image, in the testing process, we resize the image to 960 on the long side for a single scale test and get the state-of-the-art result (F-measure is 87.1\% as shown in Tabel 1).

\textbf{MSRA-TD500:} TD500 contains both Chinese and English text, and annotation boxes are line-level annotations. The blank areas between words are often included in text boxes. So instead of finetuning on IC17 pre-trained model, we train the TD500 separately, which enables the generator to generate text score maps in line form. When testing, the long side of the testing images are resized to 1600 for a single scale test. The results are shown in Table 4.

In the data sets above, IC13 and IC15 contain only English texts. The IC17 and TD500 datasets contain text in multiple languages.  Experimental results show that our algorithm has good detection effect for the multi-language, multi-rotation angle, different length, and text arrangement.

Compared with these two-stage detectors, the semantic segmentation based detectors do not train additional classifiers to precisely filter the obtained text areas, so some noise will be introduced into the detection results. Our detection results may contain some noises in order to retain some smaller characters. Fig.7 shows some failure cases.

\begin{figure}[!t]
\centering
\includegraphics[scale=0.08]{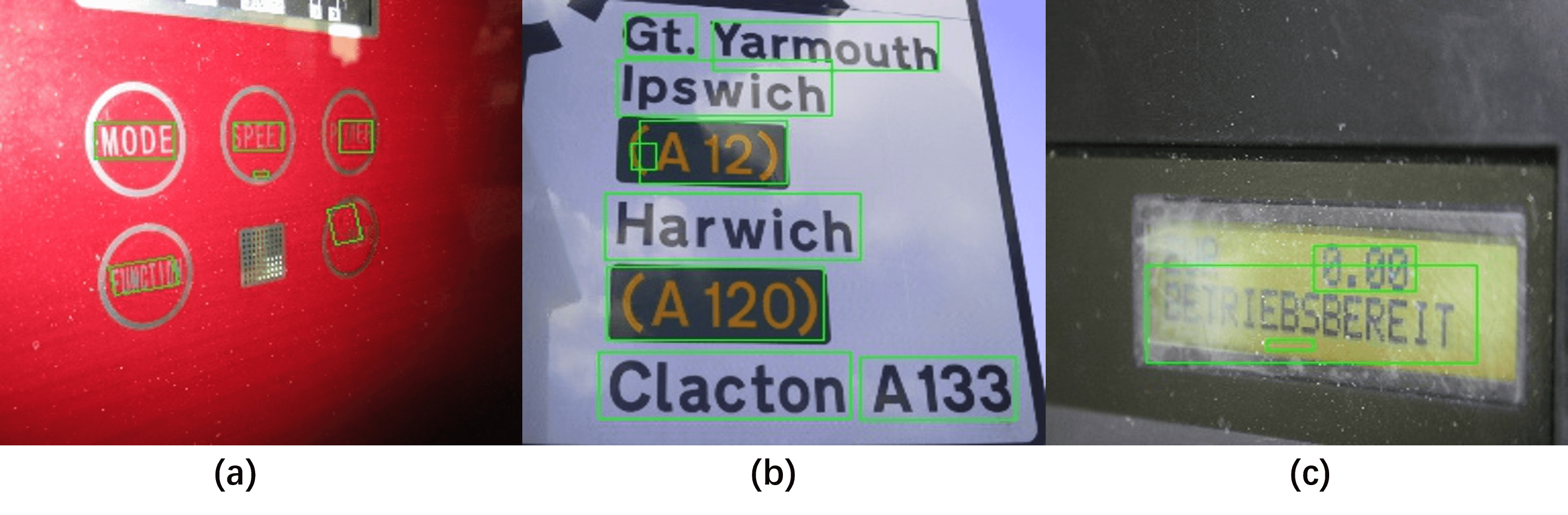}
\caption{Some failure cases of the proposed method.}
\end{figure} 

\begin{figure*}[!t]
\centering
\includegraphics[scale=0.45]{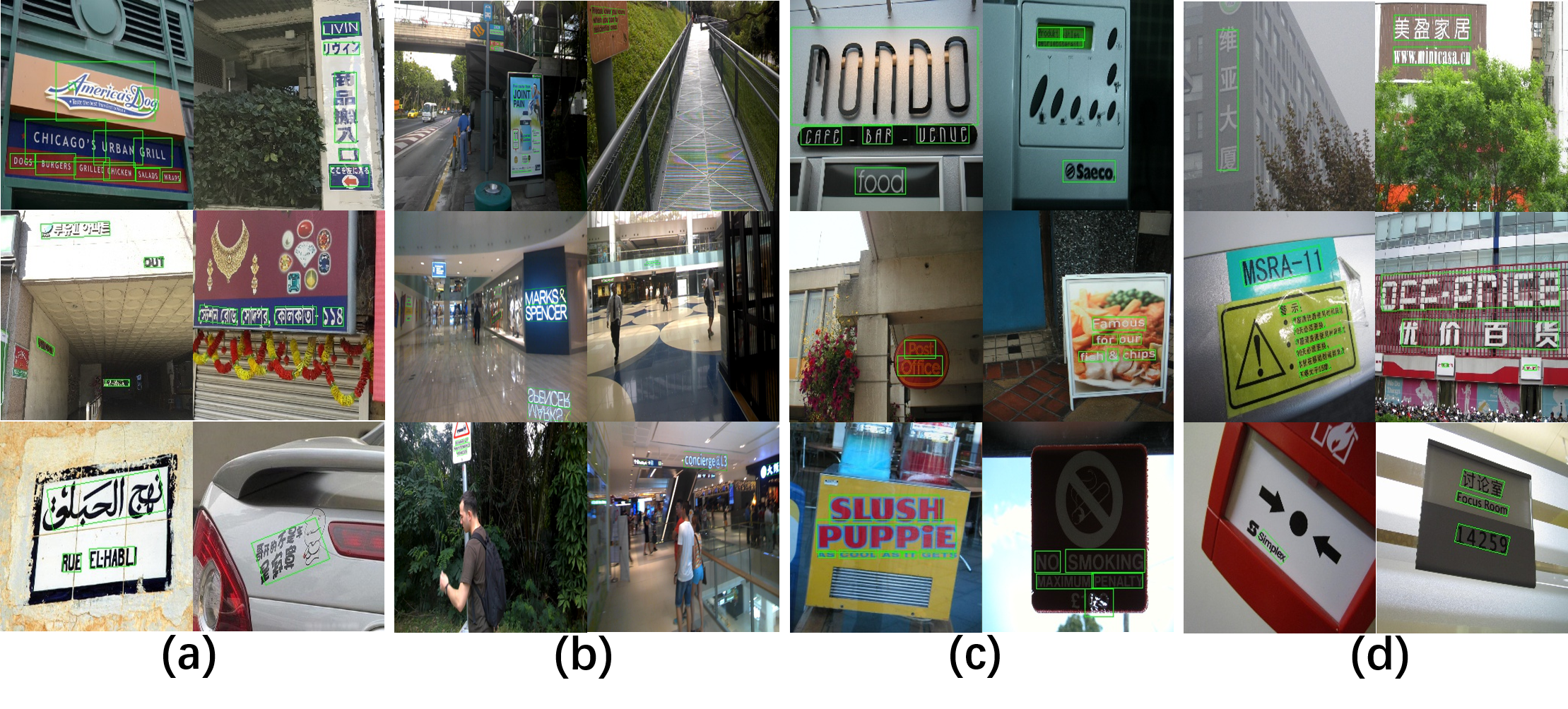}
\caption{Qualitative results of the proposed algorithm. (a) ICDAR 2017. (b) ICDAR 2015. (c) ICDAR 2013. (d) MSRA-TD500.}
\end{figure*} 

Fig.8 shows some detection results of the proposed DGST. It can be seen that the proposed method achieves potential detection results for text detection tasks in different scenarios. It has good robustness to different illumination, background and scale change, and can detect Chinese and English words effectively. At the same time, because our detector is based on the classification of pixel level, it has anti-interference to tilted and deformed text. This is also illustrated in Fig.5.

%\begin{acknowledgements}
%\end{acknowledgements}


% etc
%\end{thebibliography}
\section{Conclusion}

In this paper, we propose a novel scene text detector, DGST, which is based on the strategy of generative adversarial networks. Considering scene text detection as a special image transformation task, we introduce the idea of game theory, regard text feature extraction network as a text score image generator, and design a discriminator to identify the generated image, so that the generator can approach the labeled image step by step. In the meantime, we optimize the design of the text score image, weaken the influence of edge pixels and avoid the learning confusion problem caused by background pixels in the annotated text boxes. The experimental results on four public datasets show that our method is effective and robust.

Possible directions for future work include: (1) Explore whether the post-processing part can be replaced by a learnable network structure to reduce the use of empirical parameters.  (2) Design an end-to-end text recognition system by combining our DGST detector and a robust text recognition system. 

\section{Acknowledgment}

This work is supported by the National Natural Science Foundation of China (NSFC) under Grant No. 71621002 and the Key Programs of the Chinese Academy of Sciences under Grant No. ZDBS-SSW-JSC003, No. ZDBS-SSW-JSC004 and No. ZDBS-SSW-JSC005.

\bibliographystyle{spbasic}      % basic style, author-year citations
\bibliography{mybibfile}   % name your BibTeX data base
%\bibliographystyle{ieeetr}
%\begin{thebibliography}{}
%\section*{References}\balance
%\bibliography{mybibfile}

\end{document}